\documentclass[runningheads]{llncs}

\usepackage{iciap}

\usepackage{iciapabbrv}

\usepackage{graphicx}
\usepackage{booktabs}
\usepackage{wrapfig}
\usepackage[english]{babel}

\usepackage[accsupp]{axessibility}  % Improves PDF readability for those with disabilities.

\usepackage{hyperref}

\usepackage{orcidlink}
\usepackage{xcolor}
\usepackage[abs]{overpic}
\newcommand{\red}[1]{\textcolor{black}{#1}}

\begin{document}

\title{Physics-Based 3D Simulation for Synthetic Data Generation and Failure Analysis in Packaging Stability Assessment} 

\titlerunning{3D Simulation for Packaging Stability Assessment}

\author{
Samuel Seligardi\inst{1}\orcidlink{0009-0007-8952-4503} \and
Pietro Musoni\inst{1}$^{,*}$\orcidlink{0000-0003-3014-5983} \and
Eleonora Iotti\inst{1}\orcidlink{0000-0001-7670-2226} \and
Gianluca Contesso\inst{2} \and
Alessandro {Dal Palù}\inst{1}\orcidlink{0000-0003-0353-158X} 
}

\authorrunning{S.~Seligardi et al.}

\institute{
Department of Mathematical, Physical and Computer Sciences\\
University of Parma, Parma, Italy\\
% \email{ samuel.seligardi@studenti.unipr.it}\\
% \email{\{pietro.musoni, eleonora.iotti, alessandro.dalpalu\}@unipr.it }
$^*$\email{pietro.musoni@unipr.it}
\and
ACMI S.p.A., Fornovo di Taro, Parma, Italy\\
% \email{gianluca.contesso@acmispa.com}
}

\maketitle

\vspace*{-3ex}

\begin{abstract}
The design and analysis of pallet setups are essential for ensuring safety of packages transportation. 
With rising demands in the logistics sector, the development of automated systems utilizing advanced technologies has become increasingly crucial.
Moreover, the widespread use of plastic wrapping has motivated researchers to investigate eco-friendly alternatives that still adhere to safety standards. 
We present a fully controllable and accurate physical simulation system capable of replicating the behavior of moving pallets. It features a 3D graphics-based virtual environment that supports a wide range of configurations, including variable package layouts, different wrapping materials, and diverse dynamic conditions.
This innovative approach reduces the need for physical testing, cutting costs and environmental impact while improving measurement accuracy for analyzing pallet dynamics.
Additionally, we train a deep neural network to evaluate the rendered videos generated by our simulator, as a crash-test predictor for pallet configurations, further enhancing the system’s utility in safety analysis.
\end{abstract}

\section{Introduction}
\label{sec:intro}

In logistics safety, packages must withstand road or rail transport, safeguarding truck workers and preventing product loss or damage. 
The rigidity and strength of the load plays a critical role in transportation efficiency: permanent deformations and excessive movements of the load during transit directly impact the stability of the truck and the durability of the external wrapping, which may become deformed or torn. 

Existing safety regulations require companies to conduct evaluations on every type of product they plan to transport, ensuring compliance with stability standards throughout the transportation process.
They require specific tolerances for loaded pallets, when subjected to force caused by acceleration in a specific direction (more details in Sec. \ref{sec:standards}). 
Standardized crash tests packaging setups typically involve a physical moving platform capable of applying both positive and negative accelerations to simulate real-world transport conditions. 
It is worth noting that each package layout must be tested individually.
As a consequence, the main drawback of such a testing approach lies in its costs, in terms of energy consumption, but also in their environmental impact.

Automating the control and evaluation of these physical tests is not always straightforward. 
Regulations impose fine-grained requirements about the amount of elastic deformations of packages and the resulting condition of wrapping materials after significant acceleration changes. The analysis of a video recording of a test can detect the complete collapse of the pallet or the displacement of packages outside the pallet, by using common computer vision systems. However, current safety standards require more precise assessments and finer techniques.

Our work overcomes the limitations of the physical approach thanks to a 3D Digital Twin of the whole testing process.
The main advantages are: (i) the ability to maintain full control over packaging configurations and testing parameters; (ii) accurate measurements of all objects during the simulation and execution of large-scale experiments; (iii) our digital replica drastically reduces both economic and environmental costs associated with physical testing machinery, thus allowing packaging industries to provide larger coverage of simulated conditions and the development of optimized packaging configurations to increase safety.

\begin{figure}[tb]
    \centering
    \includegraphics[width=\columnwidth]{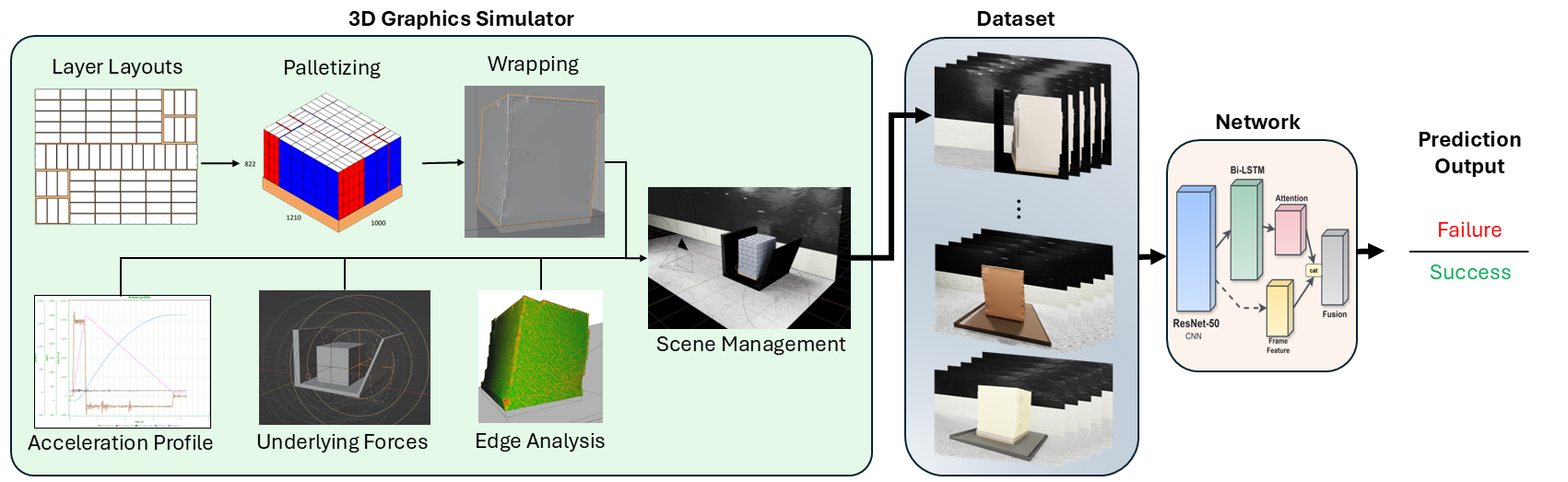}
    \caption{System's pipeline.}
    \label{fig:pipeline}
\end{figure}

Our system's pipeline, depicted in Fig.~\ref{fig:pipeline}, is composed of the following elements: a 3D graphics virtual crash test, based on rigid body dynamics combined with cloth simulation; a simulation generator (simulator) to produce synthetic renderings; an AI-powered pipeline that leverages synthetic data to perform success/failure classification and give feedback to automation engineers.

This paper presents our modeling of pallets and packages, as well as the acceleration bench, as rigid bodies subjected to forces and accelerations. 
We also represent the external envelope as a mesh with cloth dynamics (see Sec.~\ref{sec:cloth}). 
Cloth simulation ensures control over many material parameters, such as elasticity, and permits the mesh to shrink and adhere to the underlying bodies. 
Then, a simulation engine, detailed in Sec.~\ref{sec:sim}, is developed to automatically generate virtual replicas of various packaging configurations and setups, as well as acceleration conditions, with rendered videos as similar as possible to real videos obtained from the physical acceleration bench. We feed a deep neural network, detailed in Sec.~\ref{sec:exp}, that returns the classification of success or failure cases to the final users, who can adjust the configuration to improve the predicted safety levels. Sec.~\ref{sec:concl} draws the conclusions.

\vspace{-2mm}
\section{Background}
\label{sec:standards}
\label{sec:cloth}

Our virtual simulator engine aims at generating models that precisely replicate the physical testing environment. To the best of our knowledge, this is the first attempt to develop digital simulations that address safety in the transport of palletized units by focusing on their stability and rigidity under stress tests.

However, the idea of creating a 3D physical simulation to model challenging conditions is not new~\cite{felsenstein2013maritime,wang2021digital,zhao2015virtual}. Digital Twin modeling \cite{digital_twin:tao2022,pylianidis2021introducing} is one of the ways this problem is addressed.
The acceleration test bench that we model and simulate as a Digital Twin is a machinery produced by the ESTL Company \cite{estl} (from now on, the ESTL machine), designed for the packaging industry.
But, differently from conventional Digital Twins, that perfectly reproduce one single scenario, our simulator also offers a high degree of flexibility in comparison to the ESTL machine, such as changes in configurations, types of motion, and viewpoints.

Other approaches include 3D scene reconstruction, both with standard methods and deep learning \cite{guo2022neural,lyu2024total}, text-to-3D generation \cite{chen2023fantasia3d}, diffusion models \cite{liu2023meshdiffusion}, and Neural Radiance Field (NeRF) \cite{gao2022nerf}.
In our case, such methods cannot be employed due to the lack of multi-view quality data of the testing machinery. In fact, the ESTL machine cannot be modified due to safety protocols: its system provides only accelerometer data and high-speed recordings from a single fixed lateral view, capturing motion in the perpendicular direction. 

\vspace{-2mm}
\subsubsection{Safety Standards.}

\begin{figure}[t]
    \centering
    \includegraphics[width=0.45\linewidth]{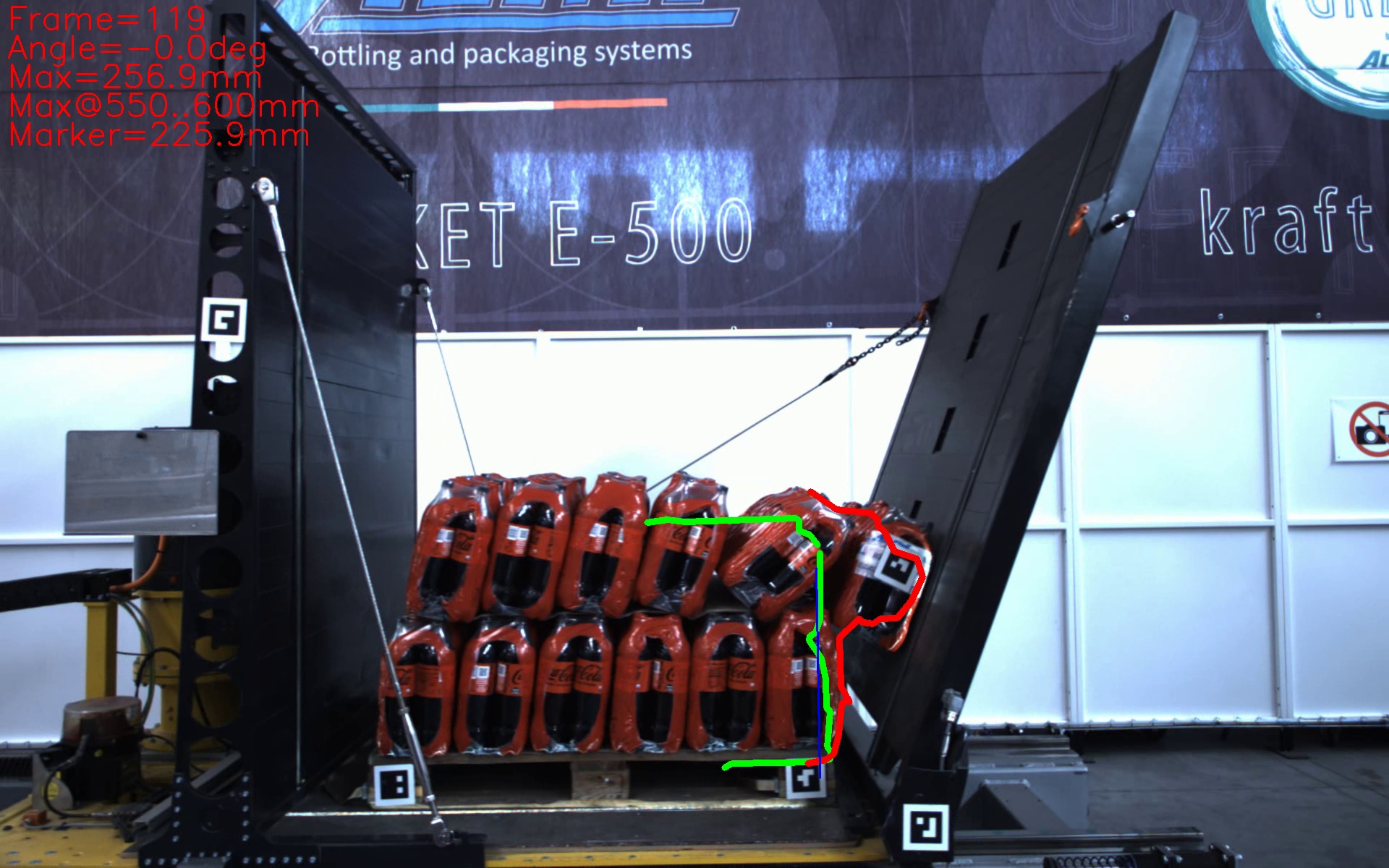}
    \includegraphics[width=0.3\linewidth]{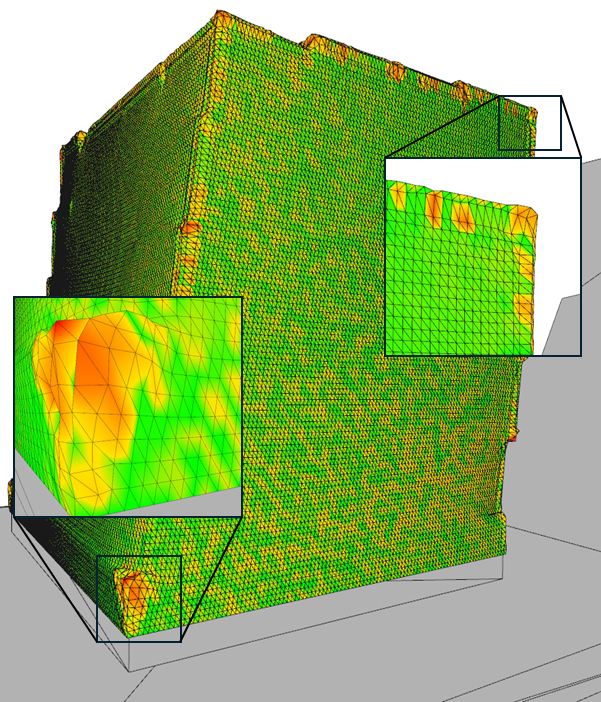}
    
    \caption{Left: Video analysis of an experimental crash-test without wrapping, showing elastic deformations (red lines) during motion. Right: Cloth simulation highlighting mesh edge stretching, with stress areas marked in red.}
    
    \label{fig:estl}
    \label{fig:edges}
  
\end{figure}

We refer to European regulations, but regulations from different countries can handled similarly by our system, having the possibility to simulate and measure any physical property involved (e.g., accelerations and displacements). The European Road Worthiness Directive \cite{euroad} introduces the basis for the EUMOS 40509 standard \cite{eumos}, which specifies how to perform palletized units crash tests and how to interpret the resulting measures as successful tests or failed tests. 
% Goal of such tests is to assess the stability of pallets when exposed to forces generated by acceleration in a defined direction.

Following EUMOS 40509, the test should provide an acceleration impulse lasting $0.5s$ followed by a constant deceleration, until the unit stops. 
The acceleration impulse to be supported is $0.5g$, but generally tests are performed in a range from $0.3g$ to $0.8g$ accelerations, in step of $0.1g$, to actually stress the loaded unit. 
Two types of deformations may occur: elastic, which are temporary and observed only during the testing window, and permanent, which represent the residual changes in the load unit after the test is completed.
A passed test requires that: (i) the permanent displacement of any part of the test load unit must not exceed $5\%$ of the total height of the load unit; (ii) within the lowest $20cm$ of the test load unit, the permanent displacement on the wooden pallet must be less than $4cm$; (iii) during testing, the elastic displacement of all parts of the test load unit must not exceed $10\%$ of its total height; (iv) no visible structural damage or product leakage should occur by the end of the test.
The ESTL machine consists of a body, called \emph{sleight}, which loads the palletized unit. The sleight moves with the tested acceleration patterns and transmits it to the loaded unit, while a high-speed recording camera records the motion. A proprietary software applies computer vision elaborations to the video and outputs the measurement of the elastic and permanent deformations, as specified by the EUMOS 40509 standard, as shown in Fig. \ref{fig:estl} on the left. 
% Accelerometers placed on the front of the sleight capture accurate accelerations on the horizontal plane during motion. 

\vspace{-2mm}
\subsubsection{Cloth Physical Simulation.} The modeling of the wrapping material is a critical aspect in simulating the behavior of a moving pallet. Some works use particle simulations \cite{shi2024robocraft}, Virtual Element Method (VEM) \cite{cihan20213d}, and neural networks \cite{malik2022hybrid} to model and simulate elasto-plastic materials. 
These approaches typically consider a single material, whereas our problem involves frequent variation in materials such as LLDPE, heat-shrink wrap, and kraft paper.
LLDPE, in various configurations of stretch, thickness, and layering, is the most used. Heat-shrink wrap has specific properties that make it more rigid than classic elasto-plastic. Kraft paper is recyclable and sustainable, but more prone to tearing. 
% More materials could be added for specific purposes. 
% Given such a great variety of parameters, reproducing the material with ad hoc simulations is undesirable, since the simulation has to be flexible.
In our work, to model the behavior of the wrapping around packages, we employ the physical simulation of cloth. Cloth simulation in animation has been a well-established research area since the early days of computer graphics. A seminal contribution in this field is the work of \cite{baraff1998large}, which laid solid groundwork for physically based cloth simulation. In recent years, the rapid growth in computational power and the development of novel techniques have significantly enhanced the realism and performance of cloth simulation systems, attracting growing interest from both researchers and industries, notably the fashion industry.
Cloth simulation in 3D animation aims to digitally approximate physical behavior and continuum properties. Various modeling approaches have been explored, including mass-spring systems \cite{kwang2002stable}, continuum-based methods \cite{wang2011data}, and yarn-level simulations \cite{kaldor2010efficient}. The parameters of these systems strongly influence the dynamics of the simulated cloth, making their selection a critical aspect of the process.
Consequently, parameter estimation and mapping between simulation parameters and real-world material properties remain active areas of research. Some 3D software platforms, such as Maya\footnote{\url{https://www.autodesk.com/it/products/maya/}} and Blender\footnote{\url{http://www.blender.org/}}, provide preset configurations for various types of fabrics. Moreover, such presets are typically only rough starting points and require manual fine-tuning.
% A recent direction in this field explores the application of deep learning techniques to cloth simulation. For instance, \cite{zhang2021deep} introduces a data-driven approach for enhancing detail. This method refines a coarse, real-time cloth simulation by learning fine-detail reconstruction from data.

\section{The Simulation Engine}
\label{sec:sim}

The core component of our pipeline is the physical simulation system (simulator).
Among many alternatives, we choose the 3D modeling software Blender, for the wide variety of customizations it enables with easy applicability, its open-source nature, and the possibility to run scripts to automatize the 3D scene generation.

Our simulation engine allows us to automate scene configuration, perform comprehensive validation checks, and render videos of packaging units subjected to various acceleration conditions.
By leveraging Blender's Python API, we created a system that automatically generates simulations of the physical behavior of various layouts of palletized loads during transport.

\subsection{Simulation System Architecture}

\begin{figure}[t]
    \centering
    \includegraphics[width=\linewidth]{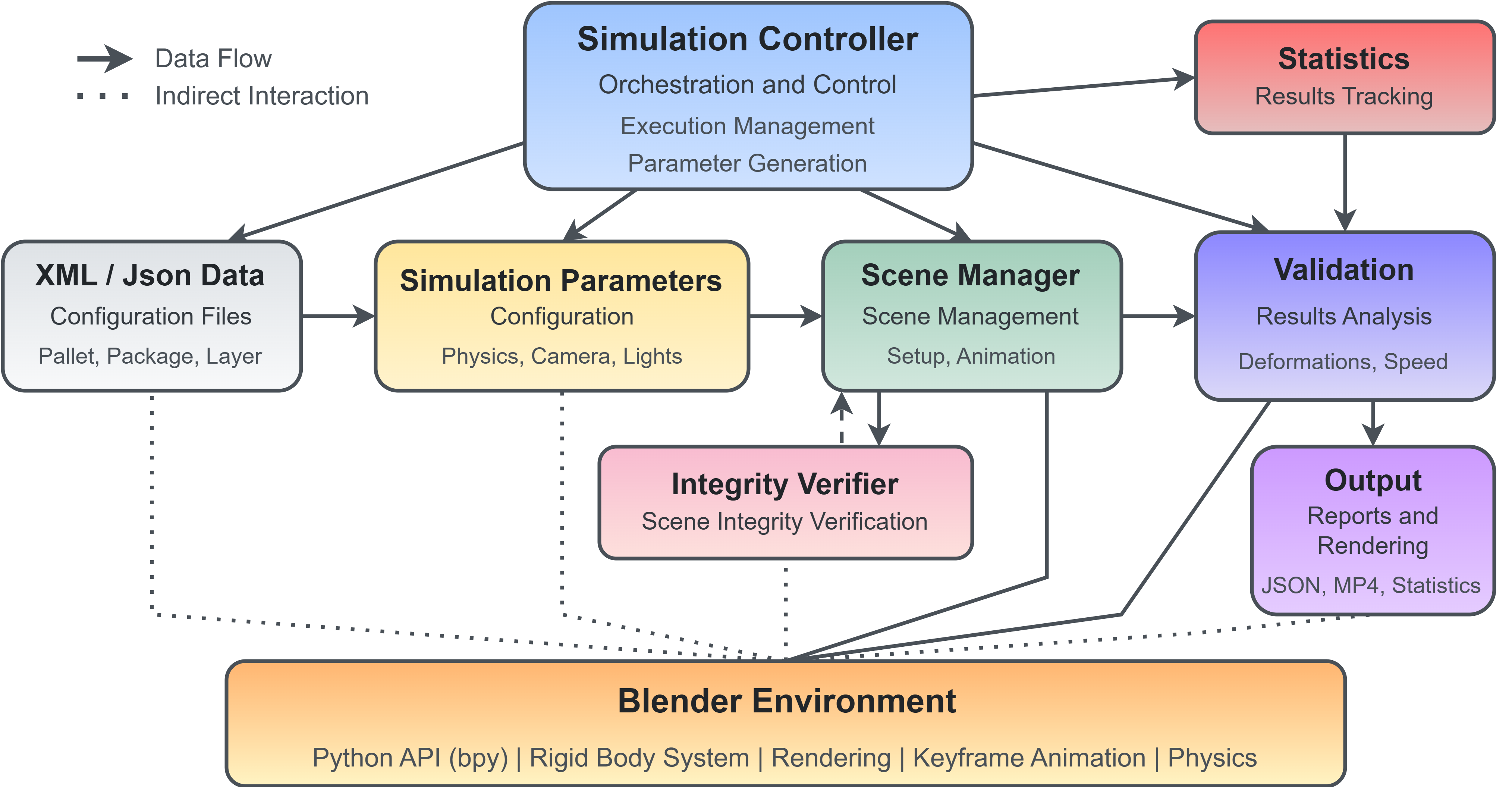}
    \caption{Architecture diagram of the simulation system showing component relationships and data flow.}
    \label{fig:architecture}
\end{figure}

Our simulator follows a modular architecture design to ensure flexibility, maintainability, and extensibility. The architecture comprises several specialized components that interact to create, execute, and validate the simulations, as shown in Fig.~\ref{fig:architecture}. 
% \red{ADP: in figura 3 non sono descritti nella legenda gli archi pieni indiretti (manca la freccia?)}

The architecture gives a clear separation of tasks, with distinct components for configuration parsing, parameter management, scene setup, physics simulation, and result analysis. Data flow from configuration files through parameter setting, scene construction, simulation execution, and finally to validation and output generation.
In detail, the system is structured around the following core components: 
\emph{the Simulation Controller} is the central component that manages execution flow, coordinates the overall simulation process, and handles parameter generation; 
\emph{the Statistics module} tracks simulation outcomes, success and failure rates, and provides reports for multiple simulation runs;
\emph{the XML Data} contains the structured configuration information including pallet specifications, package dimensions, and stacking patterns loaded from external files;
\emph{the Simulation Parameters} stores and validates all simulation parameters, including physical properties, camera settings, lighting configurations, and animation timelines;
\emph{the Scene Manager} handles all interactions with Blender environment, including object creation, material assignment, physical property configuration, and animation generation;
\emph{the Validation} performs deformation analysis and speed monitoring to determine if a simulation represents a success or failure case according to the EUMOS 40509 standard;
\emph{the Integrity Verifier} ensures that all essential objects and their properties are correctly configured before and during simulation execution, preventing potentially invalid simulations;
\emph{the Output module} manages the generation of simulation results, including rendered videos, JSON reports, and statistical data;
\emph{the Blender Environment} is the underlying 3D platform providing physics simulation, rendering capabilities, and Python API for automation. The environment comes pre-configured with essential scene objects (ESTL machine, pallet, wrapping mesh, force fields, camera, package sample), which can be modified by script to adapt for different configurations.

\subsection{Physical Modeling}

% The accuracy of our simulation system depends on its ability to model the complex physical interactions between packages, the pallet, and restraining forces. 
We employ a combination of rigid body dynamics, force fields, and cloth simulation to replicate real-world behaviors in our simulator. The fusion of these three aspects for the 3D industrial simulation in this scenario is an important contribution of our work.

\vspace{-2mm}
\subsubsection{Rigid Body Dynamics.}

The core of our physical simulation utilizes Blender's rigid body system to model packages and pallets.
The \emph{pallet} and each \emph{package} are defined as a discrete rigid body with box collision shape. 
The $i^{th}$-package $P_i$ (similarly, the pallet $P$), has the following properties. 
\begin{itemize}
    \item Dimensions extracted from configuration files: $(d_x^{(i)}, d_y^{(i)}, d_z^{(i)})$;
    \item Accurate mass values that influence physical behavior: $m_i$;
    \item Collision boundaries that define interaction surfaces: $[l_i, u_i]$;
    \item Friction coefficients that replicate surface interactions: $\varphi_i$.
\end{itemize}

The rigid body system calculates the motion of packages under external forces while handling collisions between adjacent packages.
By the control scene system, we can access the $i^{th}$-package center of mass $(x_i, y_i, z_i)$. This allows us to observe, and gather data, on how packages shift, tilt, or fall during acceleration and deceleration phases.

\vspace{-2mm}
\subsubsection{Perimeter Force Fields.}

One of the key innovations in our simulation approach is the use of perimeter force fields to replicate the restraining effect of wrapping materials. Rather than directly modeling the material's physical properties, we simulate its effect through strategically positioned force fields around the cargo.
The $k^{th}$-\emph{force fields} $F_k$, with $k$ from $1$ to $4$, are configured with the following characteristics:
\begin{itemize}
    \item Strength values proportional to cargo mass and acceleration: $F_v^{(k)}$;
    \item Positioned at a calculated distance from the cargo perimeter: $d_k$;
    \item Height-dependent influence that matches packaging behavior: $\sigma_h$;
    \item Appropriate falloff patterns to simulate realistic restraint 
    % \red{eh? che cos'è questo?}
\end{itemize}

In particular, the strength value $F_v^{(k)}$ of the $k^{th}$-force field is distributed over the cargo height: $F_v^{(k)} = F_{\text{base}}^{(k)} \cdot \sigma_h$.
The punctual strength $F_{\text{base}}^{(k)}$ is obtained from the chosen acceleration $a$ times the total mass $\Sigma_{m_i}$ of the palletized unit, also considering the part of the tension coefficient $T$ impacting on the force field:
$
F_{\text{base}}^{(k)} = \frac{ T }{4} \cdot \Sigma_{m_i} \cdot a.
$
Both $T$ and $\sigma_h$ are calibrated parameters that adjust the restraining force based on empirical observations of packaging behavior. 

\vspace{-2mm}
\subsubsection{Cloth Simulation for Wrapping Representation.}

\iffalse
\begin{figure}[tb]
  \centering
  \includegraphics[height=4.5cm]{Images/pallet_img_failure2.png}
  \caption{By using cloth simulation, we can evaluate how each edge of the mesh stretches, allowing us to identify specific points where the wrapping is under stress. In this particular case, the measurements indicate that the configuration fails the test, with clear visual evidence of stress points in the wrapping. \red{aggiungere colorbar e caso no failure}}
  \label{fig:edges}
\end{figure}
\fi

While the force fields provide the physical effects of the material of the external envelope, representing the wrapping is important for both analytical and visual presentation purposes. We use Blender's cloth simulation system to create the wrapping material that envelops the pallet load. The calculated physics of the cloth simulation allow us to manage mesh edges and monitor the critical tear points in the wrapping. As shown in Fig.~\ref{fig:edges} on the right, with the cloth simulation we can visualize stress points in the wrapping material, providing additional insights for potential failures. The simulation calculates the variation of the length of each individual mesh edge during the simulation, highlighting areas where the packaging experiences the greatest tension during transportation.

In detail, the wrapping $W$, modeled as cloth: (i) is a surface mesh that dynamically adapts to the cargo geometry; (ii) responds to the movement of the packages; (iii) provides indicators of packaging stress points; (iv) enables the identification of potential tearing locations.

The combination of rigid bodies, force fields, and cloth simulation captures both the macro-level stability of the entire load and the micro-level stress patterns in the packaging material.

\subsection{Parameters and Configurability}

A key feature of our simulator is its high degree of configurability, which allows for extensive experimentation with different package arrangements, physical properties, and testing conditions. The system's parametric approach supports both testing of specific scenarios and wide analysis through randomized configurations.
The main parameter categories include \emph{Physical Parameters}, \emph{Testing Conditions}, \emph{Visual Parameters}, \emph{Validation Thresholds}.
Physical Parameters refer to package dimensions, weights, pallet characteristics, and stacking patterns. Stacking patterns are often called \emph{palletizing schema}, and  consist of one or more layers of packages, each of these layers can be disposed with different layouts (relative positions). Layers' layouts give the major contribution to pallets stability, and our results empirically confirm this insight.
\red{The proposed setups (layouts, packages dimensions, number of layers) are real-world setups provided by manufacturing companies. The manufacturing companies use softwares that help and guide the creation of configurations, and outputs the physical parameters as XML files. Our system supports these XML files as inputs in the simulator architecture, as shown in Fig.~\ref{fig:architecture}.}
Testing Conditions consist of acceleration values ($0.3g$ to $0.8g$), acceleration duration ($0.35s$ to $0.5s$), and deceleration rates, while Visual Parameters include camera settings, lighting configurations, and rendering options.
Finally, Validation Thresholds are defined by elastic deformation limits, permanent deformation tolerances, and bottom layer specific constraints.
\begin{figure}[t]
    \centering
    \begin{subfigure}{0.4\textwidth}
        \includegraphics[width=\linewidth]{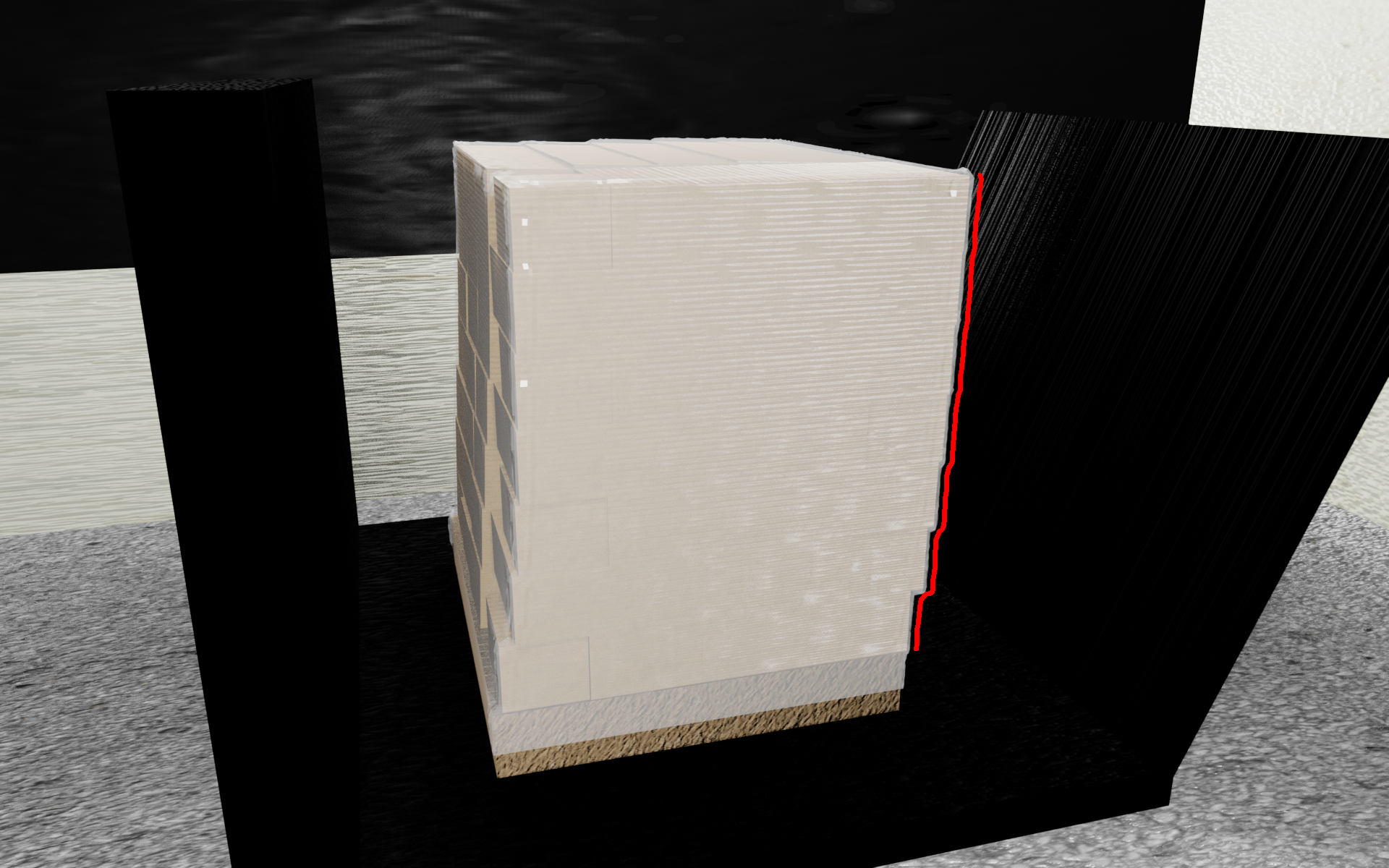}
        % \caption{Stable}
        % \caption{Stable configuration with lower acceleration ($0.5g$).}
    \end{subfigure}
    \begin{subfigure}{0.4\textwidth}
        \includegraphics[width=\linewidth]{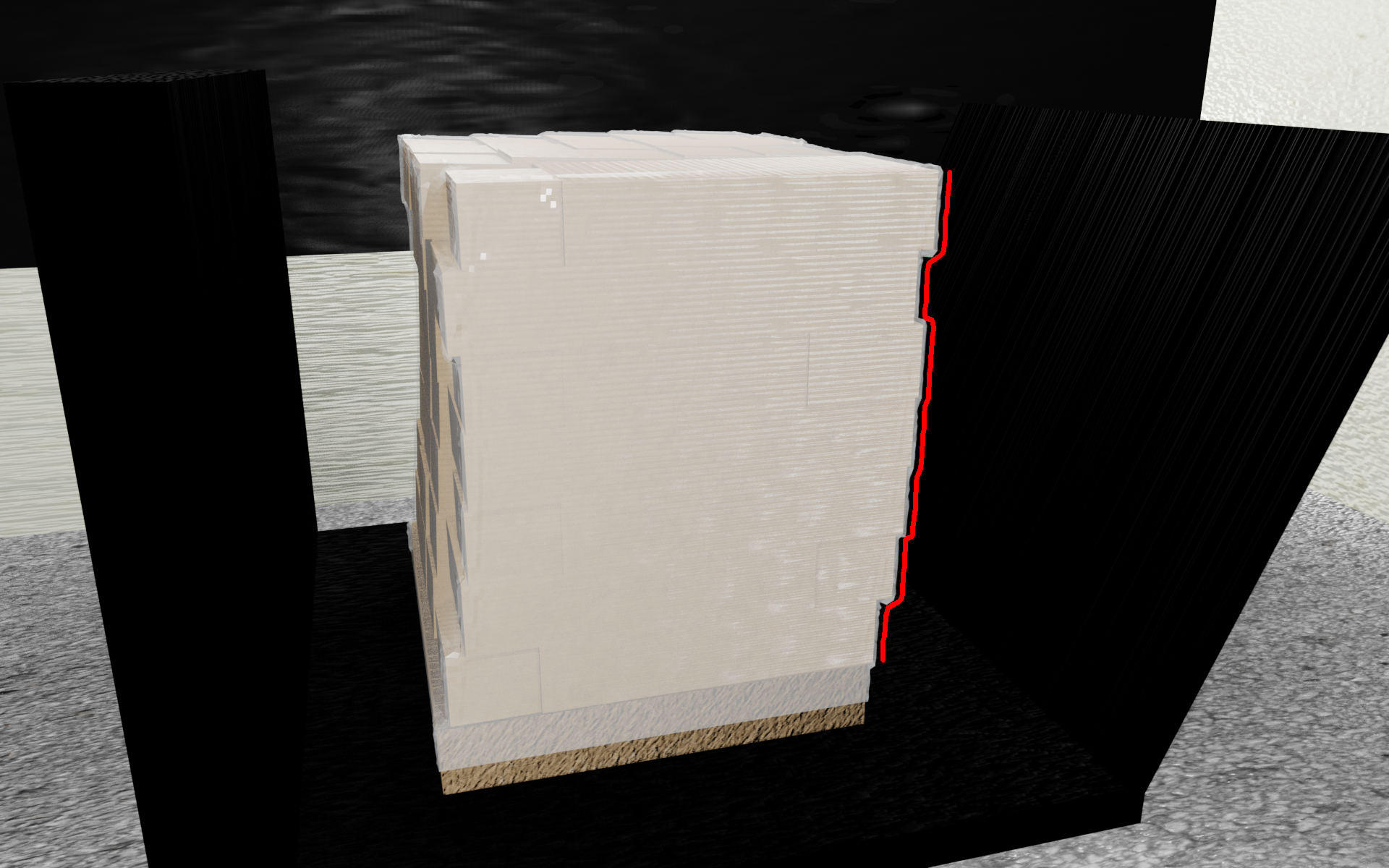}
        % \caption{Unstable}
        % \caption{Unstable configuration with higher acceleration ($0.6g$).}
    \end{subfigure}
    \begin{subfigure}{0.4\textwidth}
        \includegraphics[width=\linewidth]{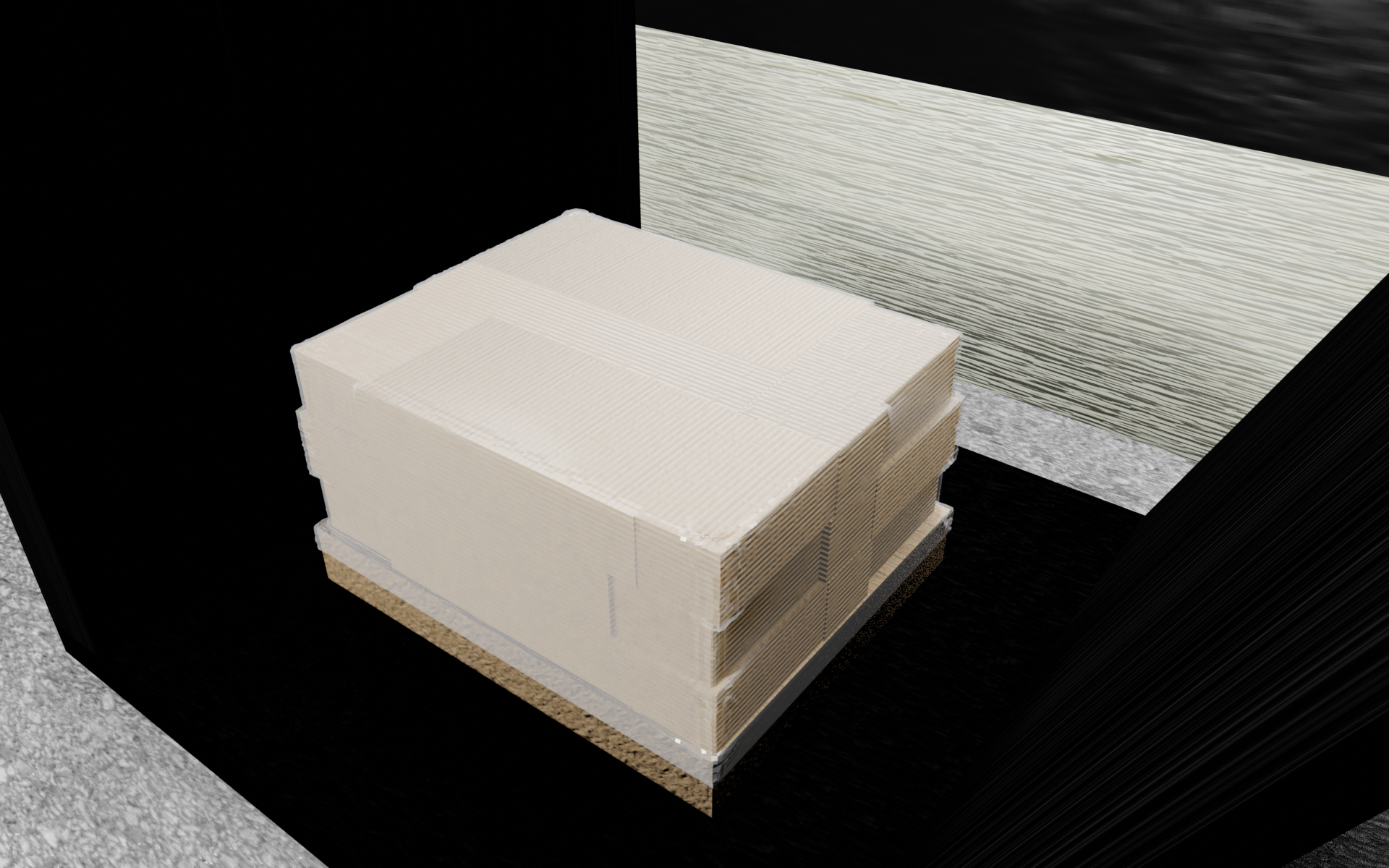}
        % \caption{Stable}
        % \caption{First pattern providing better stability during lateral acceleration.}
    \end{subfigure}
    \begin{subfigure}{0.4\textwidth}
        \includegraphics[width=\linewidth]{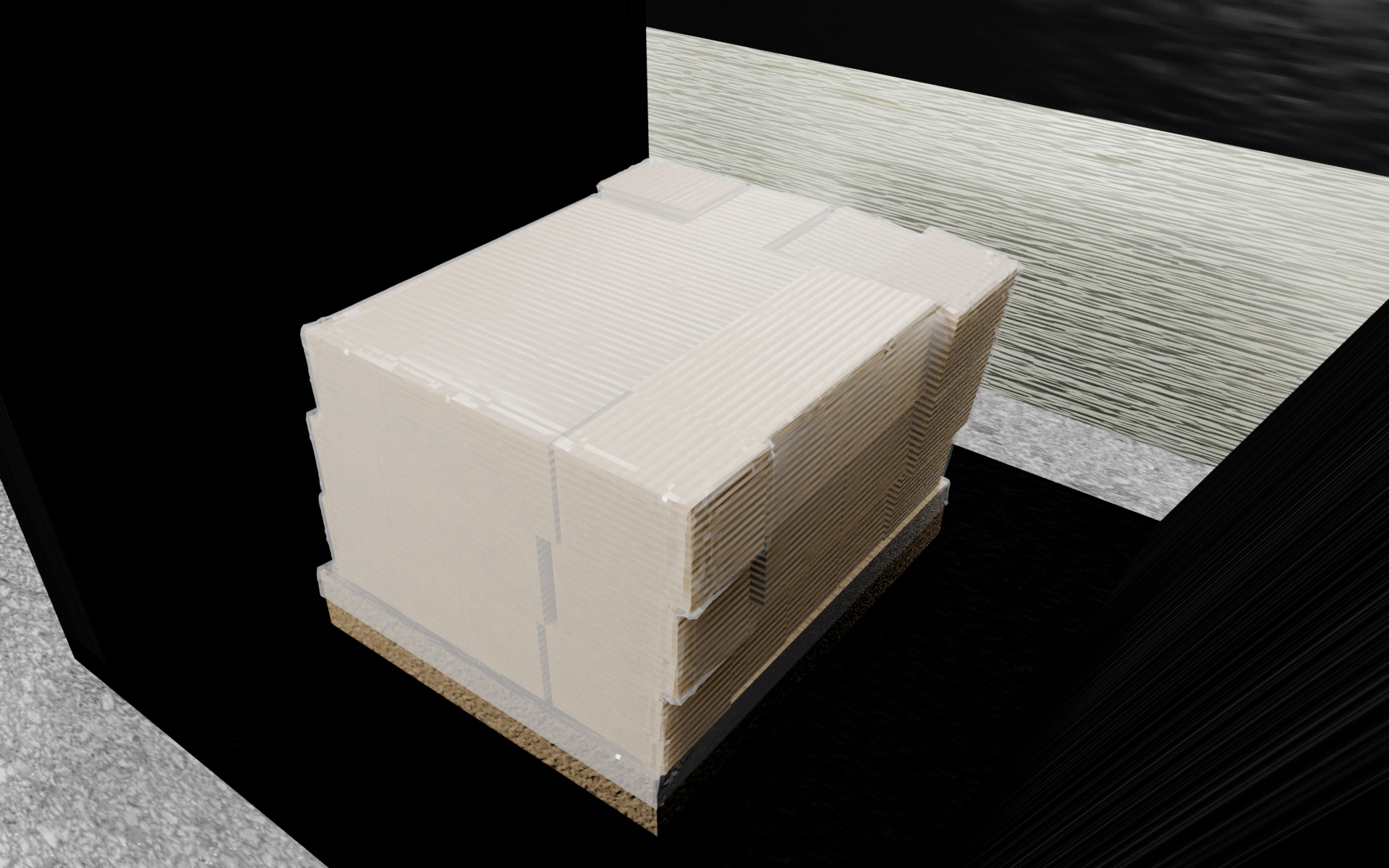}
        % \caption{Unstable}
        % \caption{Second pattern showing vulnerability to sideways forces.}
    \end{subfigure}
    \caption{Top row: Comparison of the same palletizing scheme under varying acceleration conditions. Bottom row: Different stacking patterns demonstrate how palletizing design impacts unit stability under identical test conditions. Left examples are successful; right ones show failures.}
    \label{fig:config_comparison}
\end{figure}
This comprehensive parameterization enables our simulator to generate a diverse range of scenarios. As shown in Fig.~\ref{fig:config_comparison}, even small variations in parameters can yield to significantly different outcomes, highlighting the importance of thorough testing across the parameter space.

Our implementation allows for two complementary approaches to parameter selection, i.e., \emph{deterministic configuration} and \emph{random parameter generation}.
Using the deterministic configuration approach, parameters are loaded from predefined JSON files for precise reproduction of specific test cases, useful for validation against physical tests or investigating known problematic scenarios.
Usually, the random parameter approach could be used to automatically create parameter sets within valid ranges, in order to perform massive generation, and to enable broad exploration of the parameter space, discovering unforeseen failure patterns and edge cases. The second row of Fig.~\ref{fig:config_comparison} illustrates how different palletizing patterns, one of many configurable aspects, can significantly impact load stability under identical acceleration conditions. This variability allows packaging engineers to identify optimal stacking configurations before physical implementation.

% \begin{figure}[t]
%     \centering
%     \begin{subfigure}{0.48\textwidth}
%         \includegraphics[width=\linewidth]{Images/same_acc_1.png}
%         \caption{First pattern providing better stability during lateral acceleration.}
%     \end{subfigure}
%     \hfill
%     \begin{subfigure}{0.48\textwidth}
%         \includegraphics[width=\linewidth]{Images/same_acc_2.png}
%         \caption{Second pattern showing vulnerability to sideways forces.}
%     \end{subfigure}
%     \caption{Different stacking patterns demonstrating how palletizing configuration affects unit stability under identical testing conditions.}
%     \label{fig:stacking_patterns}
% \end{figure}
\section{Experiments}
\label{sec:exp}

\begin{figure}[tb]
  \centering
  \includegraphics[width=0.9\textwidth]{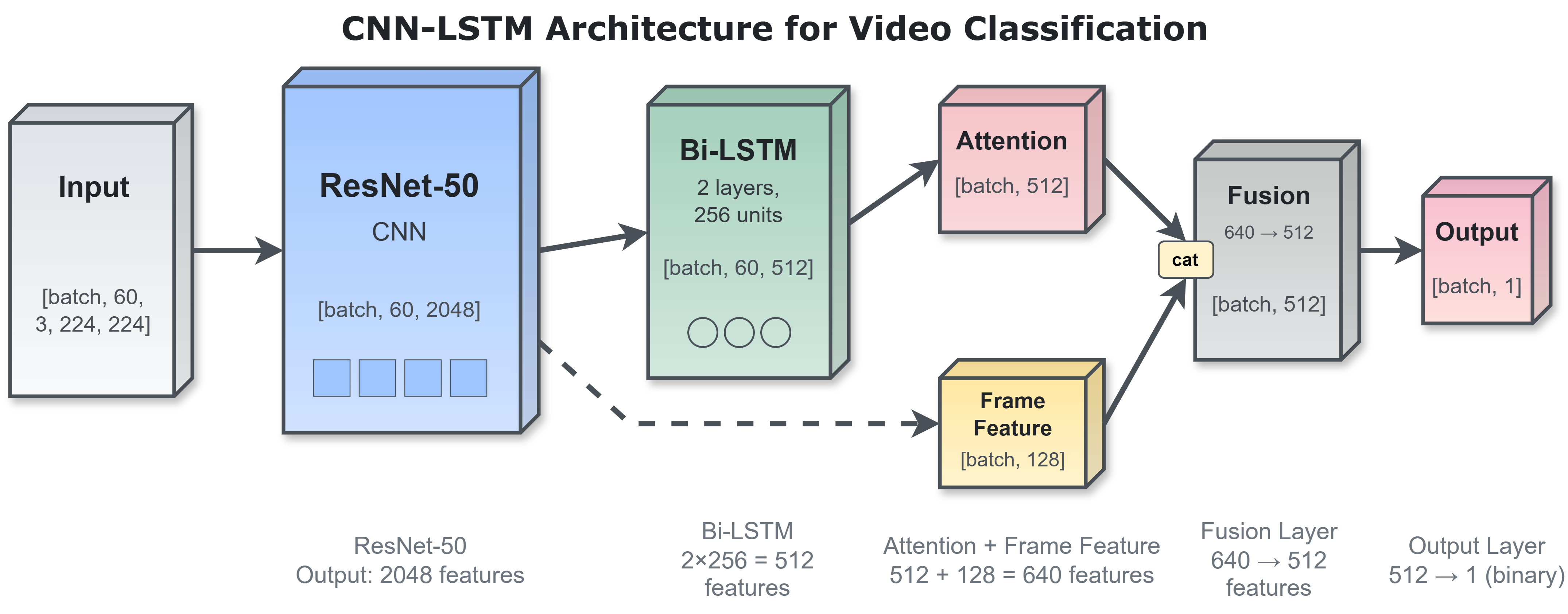}
  \caption{The architecture of the deep neural network to perform video classification.}
  \label{fig:network}
\end{figure}

\subsubsection*{Dataset.}

To evaluate the usability of our simulator in a real-world context, we generated a dataset $D = \{(v_j, o_j)\}$ of labeled videos, by rendering a diverse set of simulations across various configurations.
From real palletizing schema provided by the packaging company's software, in the form of XML configurations, we obtained package dimensions and layers layouts. The simulator varied these configurations by rotating layouts, and expanding the number of layers from $4$ to $N$, where $N$ is the original number of layers of the palletized schema.
Then, the simulator tests these augmented palletized units from $0.3g$ to $0.8g$ accelerations in step of $0.1g$, and with acceleration impulses lasting from $0.35s$ to $0.5s$.
Other variations include changing the camera position and its focal length.
The final number of rendered simulations is $1500$.
Each video $v_j$ consists of about $60 \div 70$ frames, with $1920 \times 1080$ resolution.
The simulator report for $v_j$ metadata is used for labeling, where $o_j \in \{\text{success}, \text{failure}\}$, but also carries other relevant information.
For example, in case of failure, it provides details on where the violation (whether permanent or elastic deformation) occurred. These details are used to preprocess input data to enhance elaboration.
About $1/3$ of these videos represent successful tests, while the others $2/3$ are failed tests.

This dataset is then split in $D_{\text{train}}$ and $D_{\text{test}}$ to train and validate a deep neural network classifier, capable of analyzing videos and predicting potential failure cases in virtual crash tests.
% Additionally, we validate our model by testing it on real videos of pallet movement simulations captured by ESTL machine's high-speed recordings. 
% Since training neural networks for this kind of task requires vast amounts of data, our simulator plays a crucial role by automatically generating the necessary data at scale to meet these demands.

\vspace{-2mm}
\subsubsection*{Architecture of the Network.}

\red{The network architecture is inspired by most used and efficient models in literature, such as \cite{Ge2019}.}
The architecture of the network, illustrated in Fig. \ref{fig:network}, is as follows.
Our method starts with \emph{input reshaping} to process each video frame individually using a Convolutional Neural Network (CNN). The video input, consisting of \(N_f\) frames, each a 3-channel color image at \(1920 \times 1080\) resolution, is reshaped depending on the batch size which is multiplied by the number of frames. 
Additionally, the resolution is reduced to \(224 \times 224\).
% from \((\texttt{batch\_size}, N_f, 3, 1920, 1080)\) to \((\texttt{batch\_size} \times N_f, 3, 224, 224)\), reducing the resolution to \(224 \times 224\).
Next, during \emph{feature extraction}, the ResNet-50 model processes the reshaped input, producing $2048$ features. These are leveraged for temporal sequence analysis by a Long Short-Term Memory (LSTM) model.
In \emph{temporal sequence analysis}, the LSTM processes the input feature vector, combining 256 units from each direction in its bi-directional setup, obtaining $512$ output features.
An \emph{attention mechanism} is then applied to the LSTM output, which summarizes the sequence into a single representative vector.
For \emph{frame feature fusion}, a parallel pathway is used to incorporate critical frame information. The $2048$ critical frame features are reduced to \(128\) and concatenated with the attention context vector, forming $512 + 128 = 640$ features. A fusion layer refines this to 512.
Finally, in the \emph{output prediction stage}, the fused features are passed through an output layer to generate predictions, representing the desired outcome based on the video data.

\vspace{-2mm}
\subsubsection*{Results.}

For the training process, we used different hyperparameters and optimization techniques to ensure efficient learning and prevent overfitting.
The chosen \emph{loss function} is a weighted Binary Cross-Entropy Loss, where successful test cases have a higher weight to counterbalance the ratio $1:2$ between success and failure labels.
The \emph{optimizer} is AdamW, with a learning rate of $5\cdot 10^{-4}$ and weight decay of $10^{-5}$ for better regularization.
The model was trained for a maximum of 100 epochs with an early stopping patience of 10 epochs to prevent overfitting.
We used a batch size of 8 with gradient accumulation steps of 2, effectively simulating a batch size of 16 while maintaining memory efficiency.
Automatic mixed precision training was enabled to accelerate the training process.
To further augment data, random transformations including resizing, color jitter, horizontal flips, and rotations were applied with a probability of 0.5 during training.

Our experiments demonstrate the effectiveness of the CNN-LSTM architecture for classifying pallet simulation videos. The model achieved its best performance at epoch 22, with Validation Loss of {\bf 0.2296}, Accuracy of {\bf 92.66\%}, Precision of {\bf 86.36\%}, Recall of {\bf 95.00\%}, F1 Score of {\bf 90.48\%}.
% The training progression, shown in Figure~\ref{fig:training_curves}, illustrates how the model \textit{generally improved} in its classification capability over the training period, despite some fluctuations.

% \begin{figure}[tb]
%   \centering
%   \includegraphics[width=0.8\linewidth]{Images/T1_metrics.png}
%   \caption{Training progression showing training \& validation loss and accuracy.}
%   \label{fig:training_curves}
% \end{figure}

\section{Conclusions and Future Work}
\label{sec:concl}

\begin{figure}
    \centering
    \includegraphics[width=0.32\linewidth]{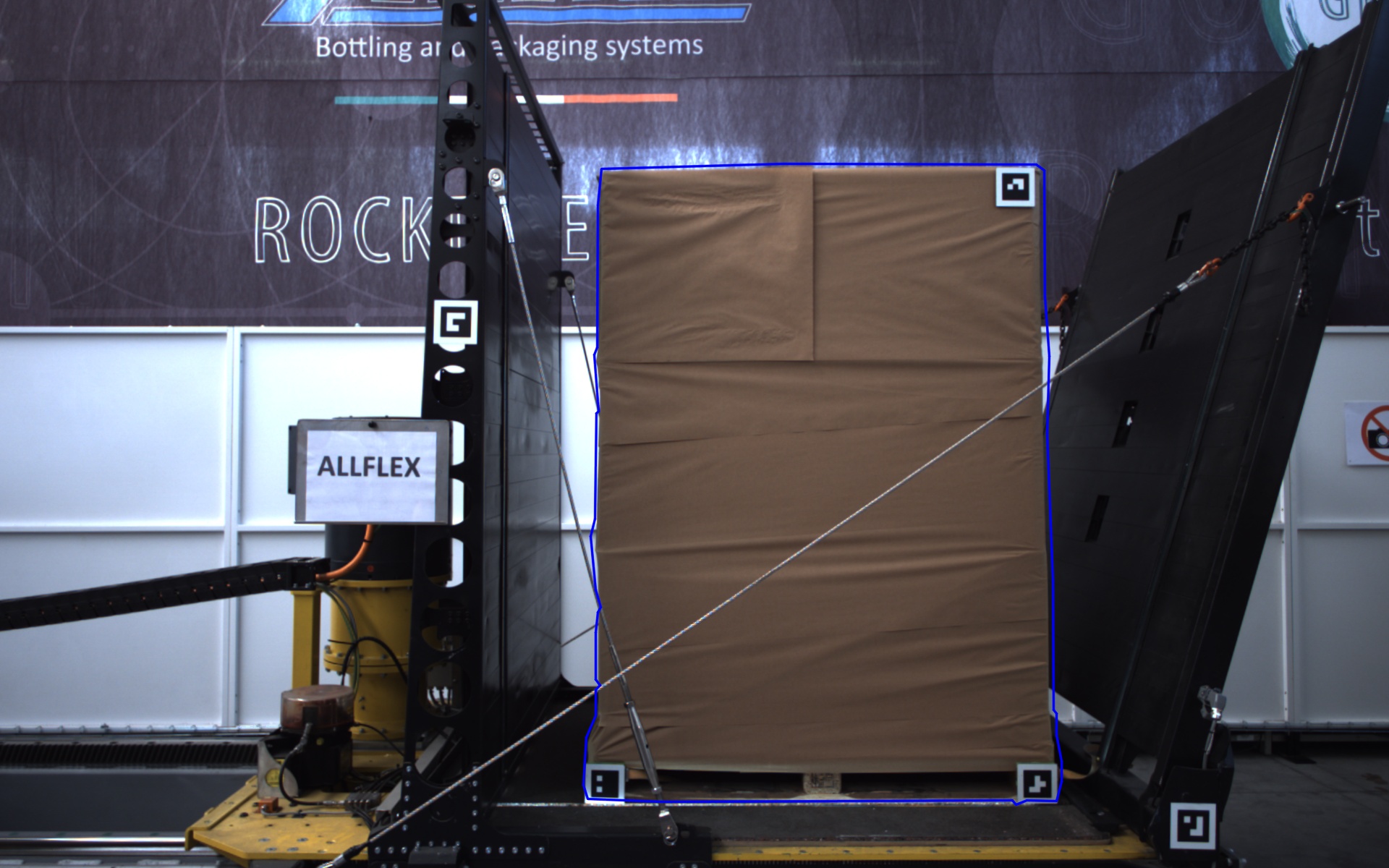}
    \includegraphics[width=0.32\linewidth]{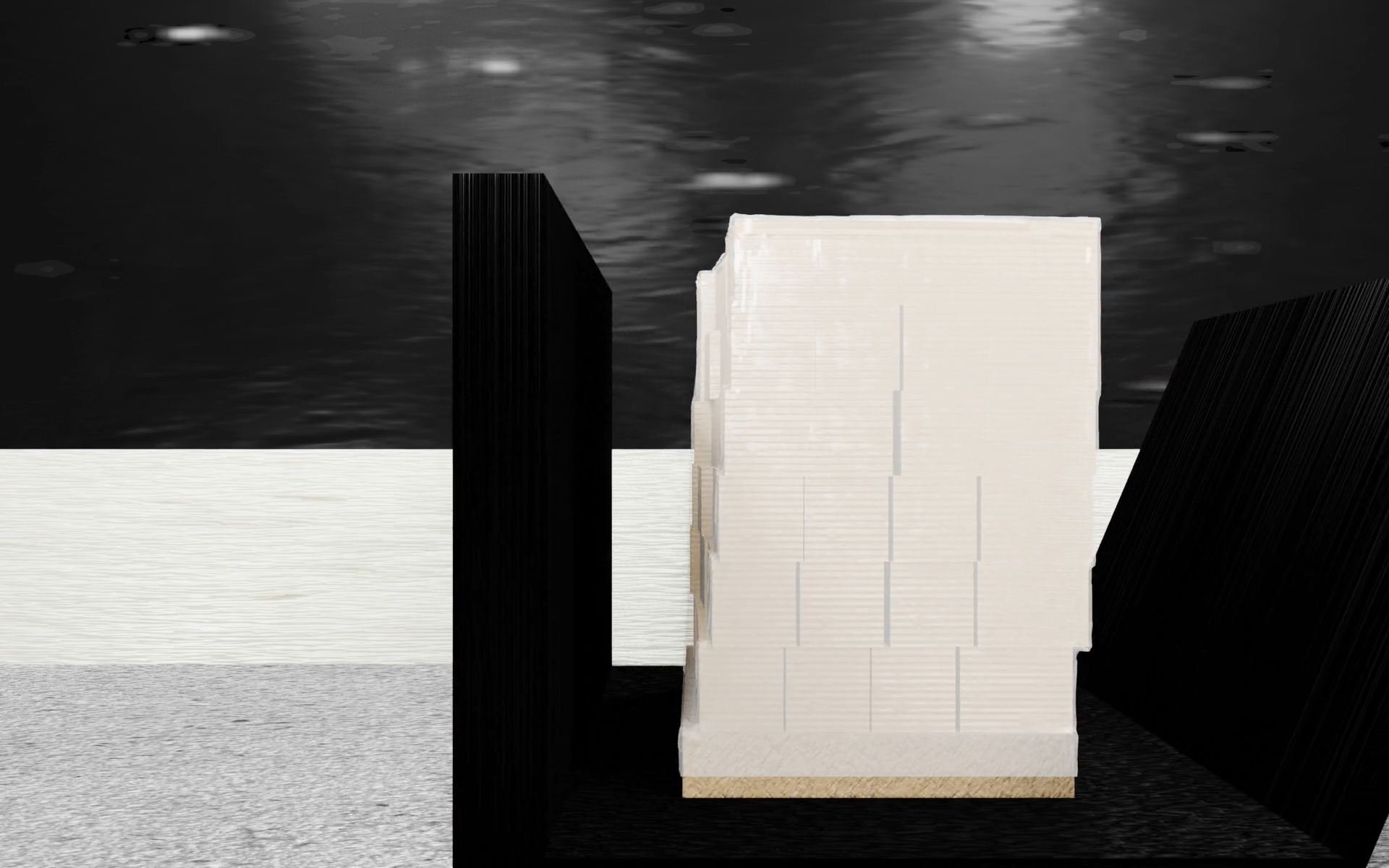}
    \includegraphics[width=0.32\linewidth]{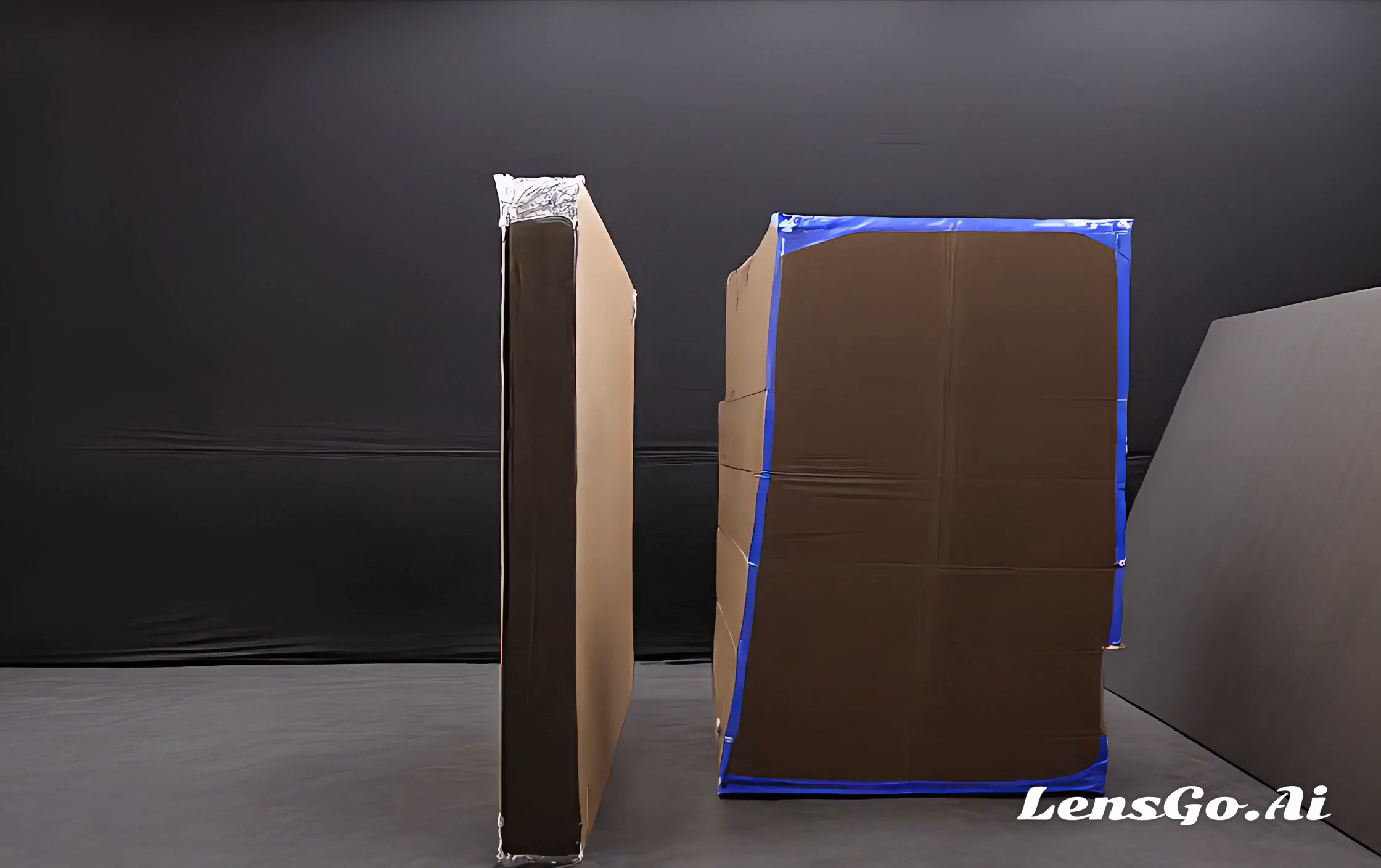}
    \caption{An example of style transfer of our simulation. First image is the style reference, a real test with paper wrapping, second image is a frame of the simulation (a configuration with plastic wrapping), third image the result created with \url{https://lensgo.ai/}.}
    \label{fig:style}
\end{figure}

In this paper, we presented the design and development of a novel virtual simulator for  crash tests of load units, based on 3D graphics of cloth. The simulator was proved capable of generating a wide variety of synthetic renderings that adhere to the real physical simulation with the acceleration bench machinery. The obtained renderings are fully controllable, and the automatically calculated metadata give precious information about test results. Moreover, the flexibility of such an instrument provides us with the ability to create different configurations, enriching the typologies of crash tests. Finally, the pipeline was completed by training a neural network that classifies success and failure cases, allowing for a complete labeling of test videos.
All European packaging companies are required to adhere to EUMOS 40509, and have great interest in cutting costs and environmental impact of these crash tests. Our simulator is the first proposal targeted at the industry world that deals with pallet stability and safety in a completely virtual 3D environment.

As observed during simulations, the layout of the packages on the pallet and the positioning of each layer significantly affect the overall stability of the load. As future development, our tool could be integrated in an automatic system to support packaging optimization and transportation planning during pallet assembly.
Our system does not allow the elastic forces of the cloth simulation to influence the motion of rigid bodies. A more flexible tool, such as NVIDIA Warp, \red{PhysX, or Bullet, }would allow us to implement this feature from scratch, improving the controllability of our system and bringing the simulation closer to real-world cloth behavior. \red{ Moreover, conditions like tearing or high strain of the external envelope could be treated separately, by measuring the tension between edges of the mesh, and detecting critical point and implemented with this tools.}
Finally, despite the rendered results of our simulations show high fidelity to the real physical behavior of these systems, the application shown by training a neural network for real-world test predictions depends on the visual similarity between the simulation videos and environmental setups. In the future, we aim to explore generative models for style transfer, which would allow us to modify our renders to better match the visual characteristics of real-world videos. A preliminary result can be seen in Fig. \ref{fig:style}. This approach opens to new challenges in style transfer including computational costs and time, the use of open-source architectures, and better identification of pallet, machine, and package materials. This would provide full control over the physics while achieving a visual result that more closely resembles the actual footage.
\vspace{-2mm}

% \begin{itemize}
%     \item use the simulator to study the configurations of the packages on the pallet
%     \item use style GAN to improve the quality of the rendered videos
%     \item implement the simulator in a more flexible framework like Nvidia warp to addess the elastic forces of the wrapping impacting on the rigid bodies (packages)
% \end{itemize}

%\par\vfill\par

\bibliographystyle{splncs04}
\bibliography{main}
\end{document}